\documentclass[prx,twocolumn,superscriptaddress,floatfix,nopacs,nofootinbib]{revtex4-2}
\usepackage[utf8]{inputenc}

\usepackage{graphicx,amsfonts,amssymb,amsmath,hyperref,enumerate}

\usepackage{algorithm}
\usepackage{algpseudocode}
\usepackage{pgfplots}
\usepackage{pgf}
\usepackage{lmodern}
\usepackage{import}
\usepackage{xr}
\usepackage{enumitem}
\usepackage{soul}
\usepackage[normalem]{ulem}
\usepackage{multirow}
\usepackage{subcaption}

\newif\ifhyper
\hypertrue
\ifhyper
\hypersetup{
   citecolor = {green},
   colorlinks = {true}, 
   urlcolor = {blue} 
}
\fi

\newcommand{\beq}{\begin{equation}}
\newcommand{\eeq}{\end{equation}}
\newcommand{\beqa}{\begin{eqnarray}}
\newcommand{\eeqa}{\end{eqnarray}}

\newcommand{\comment}[1]{}

\def\Longarrow{\protect\@lra}
\def\@lra{\relbar\joinrel\relbar\joinrel\relbar\joinrel%
          \relbar\joinrel\rightarrow}

\pgfplotsset{compat=1.18}
\usepackage[utf8]{inputenc}

\begin{document} 

\title{Synthetic Data Generation and Differential Privacy \\ using Tensor Networks' Matrix Product States (MPS)}



\author{Alejandro Moreno R.} 
\affiliation{Multiverse Computing, Parque Cientifico y Tecnol\'{o}gico de Gipuzkua, Paseo de Miram\'{o}n, 170 3$^{\,\circ}$ Planta, 20014 Donostia / San Sebasti\'{a}n, Spain}
\author{Desale Fentaw}
\affiliation{Multiverse Computing, Parque Cientifico y Tecnol\'{o}gico de Gipuzkua, Paseo de Miram\'{o}n, 170 3$^{\,\circ}$ Planta, 20014 Donostia / San Sebasti\'{a}n, Spain}
\author{Samuel Palmer}
\author{Raul~Salles~de~Padua}
\affiliation{Multiverse Computing, Parque Cientifico y Tecnol\'{o}gico de Gipuzkua, Paseo de Miram\'{o}n, 170 3$^{\,\circ}$ Planta, 20014 Donostia / San Sebasti\'{a}n, Spain}
\author{Ninad~Dixit}
\affiliation{Multiverse Computing, Parque Cientifico y Tecnol\'{o}gico de Gipuzkua, Paseo de Miram\'{o}n, 170 3$^{\,\circ}$ Planta, 20014 Donostia / San Sebasti\'{a}n, Spain}
\author{\\Samuel Mugel}
\affiliation{Multiverse Computing, Parque Cientifico y Tecnol\'{o}gico de Gipuzkua, Paseo de Miram\'{o}n, 170 3$^{\,\circ}$ Planta, 20014 Donostia / San Sebasti\'{a}n, Spain}

\author{Roman Orus}

\affiliation{Multiverse Computing, Parque Cientifico y Tecnol\'{o}gico de Gipuzkoa, Paseo de Miram\'{o}n, 170 3$^{\,\circ}$ Planta, 20014 Donostia / San Sebasti\'{a}n, Spain}


\author{\\Manuel Radons}
\author{Josef Menter}
\author{Ali Abedi}
\affiliation{Bundesdruckerei GmbH, Kommandantenstraße 18, 10969 Berlin, Germany}

\begin{abstract}
Synthetic data generation is a key technique in modern artificial intelligence, addressing data scarcity, privacy constraints, and the need for diverse datasets in training robust models. In this work, we propose a method for generating privacy preserving high-quality synthetic tabular data using Tensor Networks, specifically Matrix Product States (MPS)~\cite{Wang2023}.We benchmark the MPS-based generative model against state-of-the-art models such as CTGAN~\cite{xu2018synthesizing}, VAE, and PrivBayes~\cite{golob2024high}, focusing on both fidelity and privacy-preserving capabilities. To ensure differential privacy (DP), we integrate noise injection and gradient clipping during training, enabling privacy guarantees via Rényi Differential Privacy accounting~\cite{giomi2022unified}. Across multiple metrics analyzing data fidelity and downstream machine learning task performance, our results show that MPS outperforms classical models, particularly under strict privacy constraints. This work highlights MPS as a promising tool for privacy-aware synthetic data generation. By combining the expressive power of tensor network representations with formal privacy mechanisms, the proposed approach offers an interpretable and scalable alternative for secure data sharing. Its structured design facilitates integration into sensitive domains where both data quality and confidentiality are critical.

\end{abstract}

\maketitle

\section{Introduction}
\label{sec:intro}

The growing demand for large, diverse, and privacy-compliant datasets has established synthetic data generation as a cornerstone of modern machine learning pipelines. In domains such as healthcare and finance, where data sensitivity is paramount, the ability to simulate realistic data while protecting individual information enables secure and scalable development of AI systems.

Despite its benefits-scalability, accessibility, and flexibility-synthetic data also poses risks. Since it is generated by models trained on real datasets, it may inadvertently leak sensitive information, raising concerns about privacy and confidentiality~\cite{Abadi_2016}. Differential Privacy (DP) has emerged as a principled solution to these challenges, offering formal guarantees by injecting calibrated noise into the data generation process. When properly implemented, DP ensures that the inclusion or exclusion of any individual in the dataset does not significantly affect the output.

Concurrently, Tensor Networks (TNs)-a class of models from quantum physics-offer promising tools for high-dimensional data representation. Among them, Matrix Product States (MPS) provide a structured and interpretable parameterization of complex joint distributions, with linear scaling in the number of features. While MPS models have shown success in representing binary data~\cite{Wang2023}, their application to mixed-type synthetic tabular data remains underexplored.

In this work, we extend the MPS framework to support categorical, integer, and continuous features, and embed differential privacy mechanisms directly into the training loop using gradient clipping and noise injection. This integration enables a strong balance between utility and formal privacy guarantees. By exploiting the structural advantages of MPS, our approach offers a transparent and scalable alternative to black-box generative models.

Recent advances in synthetic tabular data generation-such as Conditional Tabular GANs (CTGAN) and Variational Autoencoders (VAE)-have shown proficiency in capturing statistical properties and supporting downstream tasks. However, these models often involve complex architectures, require extensive hyperparameter tuning, and offer limited interpretability and native privacy support. Our MPS-based approach addresses these limitations with a compact, interpretable design and built-in privacy accounting.

The remainder of this paper is organized as follows. Section~\ref{sec:methods} reviews prior work in synthetic data generation and differential privacy. Section~\ref{sec:problem} presents the MPS model, preprocessing pipeline, and privacy integration. Section~\ref{sec:results} evaluates performance against baseline models. Section~\ref{sec:conclusion} discusses implications and limitations, and the Appendix provides extended metric breakdowns and visualizations.

\section{Related Work} 
\label{sec:methods}

\subsection{Deep Generative Models for Tabular Data}

The generation of synthetic tabular data has garnered increasing attention with the rise of data-driven applications constrained by privacy, regulatory, or availability concerns. Among the most prominent techniques are Generative Adversarial Networks (GANs)~\cite{goodfellow2014generative} and Variational Autoencoders (VAEs)~\cite{kingma2013auto}, which have shown strong empirical performance in generating realistic, high-dimensional data. However, their application to tabular domains introduces unique challenges, such as modeling both discrete and continuous variables, managing imbalanced categorical features, and preserving high-fidelity dependencies across features.

CTGAN~\cite{xu2018synthesizing}, a GAN-based framework tailored for tabular data, addresses these issues by employing conditional generation and specialized training objectives to avoid mode collapse. It has become a benchmark model due to its ability to capture complex joint distributions. However, CTGAN requires extensive architecture tuning and lacks native support for formal privacy mechanisms.

VAEs adopt a probabilistic latent-variable framework and optimize the evidence lower bound (ELBO) to approximate the data distribution. While they typically produce smooth and diverse samples, VAEs struggle with discrete and high-cardinality categorical features due to the limitations of the Gaussian latent space. Extensions such as the Tabular VAE (TVAE) attempt to mitigate this through hybrid encoding schemes, though categorical fidelity often remains inferior to adversarial models.

These models-CTGAN, VAE, and TVAE-serve as classical baselines for our evaluation.

\subsection{Privacy Mechanisms in Generative Modeling}

Differential Privacy (DP)~\cite{Dwork2006Calibrating} provides a principled framework for protecting individual data entries in statistical computations. A randomized mechanism $\mathcal{M}$ satisfies $(\epsilon, \delta)$-DP if for all measurable sets $S$ and for any pair of neighboring datasets $D$ and $D'$ differing in one element:
\begin{equation}
\Pr[\mathcal{M}(D) \in S] \leq e^\epsilon \Pr[\mathcal{M}(D') \in S] + \delta.
\end{equation}

In deep learning, DP is typically achieved by injecting noise into gradients during training, combined with gradient clipping to bound sensitivity~\cite{Abadi_2016}. Privacy loss over time is tracked using frameworks such as Rényi Differential Privacy (RDP)~\cite{Mironov2017RDP} and the moment accountant technique.

While DP-GANs and DP-VAEs have been proposed, the inherent instability of training under noise often leads to utility degradation. PrivBayes~\cite{zhang2017privbayes} adopts an alternative approach by training Bayesian networks with explicit DP constraints. It achieves strong privacy guarantees but suffers from scalability limitations in high-dimensional settings.

\subsection{Tensor Networks and Born Machines}

Tensor Networks (TNs)~\cite{orus2014practical, Verstraete_2008} originated in quantum physics as a way to efficiently represent high-dimensional entangled states. Among these, the Matrix Product State (MPS)-also known as the tensor train-factorizes a high-order tensor into a sequential chain of low-rank tensors, enabling efficient modeling of global correlations with parameters that scale linearly in the number of variables.

\begin{figure}[ht]
\centering
\includegraphics[width=0.45\textwidth]{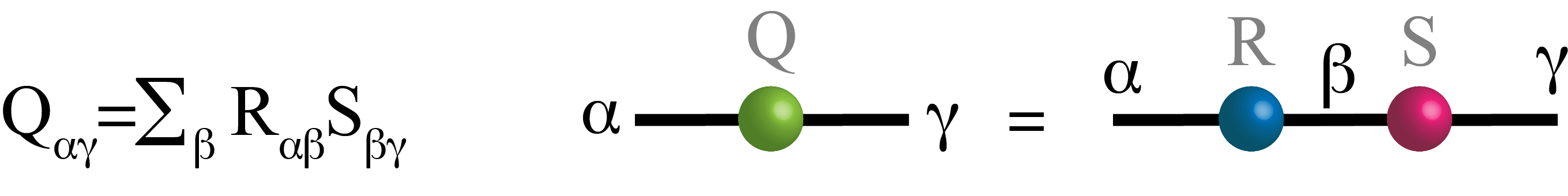}
\caption{Contraction of tensors, equivalent to matrix multiplication, is graphically represented by connecting the shared indices.}
\label{Fig:contraction}
\end{figure}

A key operation in TNs is tensor contraction. For instance, the contraction of two tensors $R_{\alpha\beta}$ and $S_{\beta\gamma}$ over the shared index $\beta$ results in:
\begin{equation}
Q_{\alpha\gamma} = \sum_\beta R_{\alpha\beta} S_{\beta\gamma},
\end{equation}
which is graphically represented in Figure~\ref{Fig:contraction}.

\begin{figure}[ht]
\centering
\includegraphics[width=0.45\textwidth]{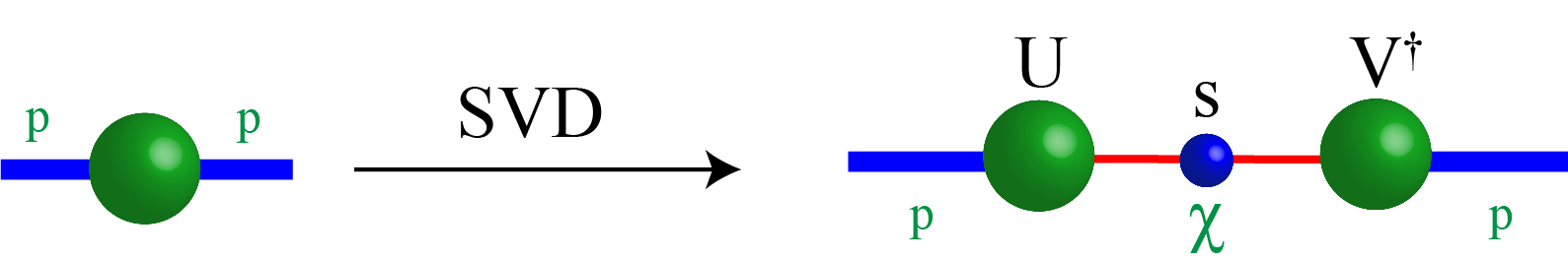}
\caption{Singular Value Decomposition (SVD) breaks a matrix into low-rank tensors for efficient representation, supporting MPS compression.}
\label{Fig:svd}
\end{figure}

The MPS structure relies heavily on Singular Value Decomposition (SVD), which enables compression by truncating insignificant singular values, thereby preserving only the dominant statistical correlations (see Figure~\ref{Fig:svd}).

\begin{figure}[h]
\centering
\begin{tikzpicture}[thick, every node/.style={scale=0.9}, tensor/.style={draw, circle, minimum size=1cm}, leg/.style={line width=0.8pt}]
\node[tensor] (A1) at (0,0) {$A^{[1]}$};
\node[tensor] (A2) at (2,0) {$A^{[2]}$};
\node[tensor] (A3) at (4,0) {$A^{[3]}$};
\node[tensor] (A4) at (6,0) {$A^{[4]}$};
\draw[leg] (A1) -- ++(0,-1) node[below] {$x_1$};
\draw[leg] (A2) -- ++(0,-1) node[below] {$x_2$};
\draw[leg] (A3) -- ++(0,-1) node[below] {$x_3$};
\draw[leg] (A4) -- ++(0,-1) node[below] {$x_4$};
\draw[leg] (A1) -- (A2);
\draw[leg] (A2) -- (A3);
\draw[leg] (A3) -- (A4);
\node at (3, 1.2) {\textbf{MPS-based Born Machine}};
\node at (3, -2) {\small $P(x_1,x_2,x_3,x_4) = \frac{|\psi(x_1,x_2,x_3,x_4)|^2}{Z}$};
\end{tikzpicture}
\caption{Born Machine with MPS: inputs $x_i$ are mapped to tensor cores $A^{[i]}$; sampling proceeds through sequential contractions.}
\label{Fig:born}
\end{figure}
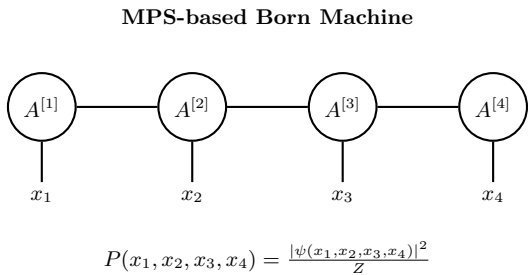

In the context of generative modeling, the Born Machine~\cite{Han2018MPSgen} defines the probability of a binary configuration $x$ as the squared amplitude of a quantum wavefunction:
\begin{equation}
P(x) = |\Psi(x)|^2, \quad \Psi = \sum_x \Psi(x)|x\rangle.
\end{equation}
The wavefunction $\Psi(x)$ is parameterized via an MPS, and exact sampling is performed through sequential contractions of the network cores (Figure~\ref{Fig:born}).

Tensor networks have demonstrated promise in machine learning applications, including supervised classification~\cite{Stoudenmire2016SupervisedLW}, clustering~\cite{Stoudenmire_2018}, anomaly detection~\cite{Wang2020, aizpurua2023}, and solving PDEs via neural networks~\cite{Patel2022}. However, their use in synthetic data generation for mixed-type tabular datasets remains limited. Moreover, privacy-preserving extensions to TN-based generative models are scarcely explored.
A notable exception is Pozas-Kerstjens et al.~\cite{pozas2024privacy}, which investigates gauge-induced privacy within MPS and neural networks. Unlike our work, which integrates privacy directly into the training algorithm using noise injection and formal DP accounting, that approach focuses on structural transformations post-training.

\section{Problem Formulation}
\label{sec:problem}


This work extends the Matrix Product State (MPS) framework, originally developed for modeling binary features, to the generation of realistic synthetic tabular data comprising of categorical, integer, and continuous attributes. Our primary objective is to construct a generative model that produces high-fidelity data reflecting real-world distributions, while rigorously enforcing Differential Privacy (DP) guarantees during training.

Integrating DP into synthetic data generation introduces a fundamental challenge: maintaining the utility of the synthetic data while safeguarding privacy. To address this, during the MPS model training we integrate stochastic gradient descent (SGD) with gradient clipping and noise injection. These mechanisms are compatible with DP accounting methods, particularly moment accounting based on Rényi Differential Privacy (RDP) \cite{Abadi_2016, Mironov_2017}, allowing us to track cumulative privacy loss and set strict guarantees on the $\epsilon$-differential privacy budget. These techniques ensure that sensitive information is obscured while minimizing performance degradation in downstream tasks.

Let $\mathcal{D}_{\text{real}}$ be the original dataset. We aim to learn a probabilistic model $P_\theta(x)$, parameterized via an MPS, that approximates the data distribution:
\[
P_\theta(x) \approx P_{\text{real}}(x), \quad \text{with } x \in \mathcal{X},
\]
such that training the model on $\mathcal{D}_{\text{real}}$ satisfies $(\epsilon, \delta)$-DP.



Our training process is designed to minimize the negative log-likelihood of the data under the MPS model while adhering to the defined DP budget. As a result, the synthetic data maintains strong alignment with the original distribution, both statistically and for downstream tasks.

In summary, this work formalizes the generation of privacy-preserving tabular synthetic data as a constrained learning problem, with the objective:
\[
\min_{\theta} \; \mathbb{E}_{x \sim \mathcal{D}_{\text{real}}}[-\log P_\theta(x)] \quad \text{subject to} \; (\epsilon, \delta)\text{-DP}.
\]

\subsection{Dataset description}
\label{sec:dataset}

To evaluate the performance of our MPS-based synthetic data generator, we use the well-established Adult Income dataset from the 1996 U.S. Census Bureau \cite{Dataset}. This dataset contains 48,842 records with a mix of 6 integer/continuous and 8 categorical features, offering a representative benchmark for mixed-type tabular data.

Approximately 7\% of the entries contain missing values. These were handled during preprocessing to ensure model consistency.

Below is a list of the attributes included in the dataset:
\begin{enumerate} 
    \item Age: Integer, from 17 to 90.
    \item Workclass: Categorical (8 categories)
    \item Fnlwgt: Integer from 12285 to 1490400.
    \item Education: Categorical (16 categories)
    \item Education-num: Integer from 1 to 16.
    \item Marital-status: Categorical (7 categories)
    \item Occupation: Categorical (14 categories)
    \item Relationship: Categorical (6 categories)
    \item Race: Categorical (5 categories)
    \item Sex: Categorical (2 categories)
    \item Capital-gain: Continuous from 0 to 99999.
    \item Capital-loss: Continuous from 0 to 4356.
    \item Hours-per-week: Integer from 1 to 99.
    \item Native-country: Categorical (41 categories)
    \item Class: Categorical (2 categories)
\end{enumerate}

The diverse feature set including highly skewed and high-cardinality variables provides a comprehensive test bed for evaluating both synthetic data fidelity and privacy performance.  

\section{Methodology}
\label{sec:experimental_setup}

This section details the preprocessing pipeline, privacy-preserving mechanisms, training procedures, and evaluation metrics used to implement and benchmark the MPS-based synthetic data generator.

\subsection{Data Preprocessing}

The first challenge is embedding tabular data features into formats compatible with the MPS model, which operates on discrete, tensor-structured inputs. The dataset is split into categorical and continuous features, with separate preprocessing for each type.

\textbf{Categorical features} are mapped into high-dimensional orthogonal bases, such as one-hot or index embeddings. Each categorical variable is represented by a single tensor core, with a physical leg indexed over its category values:
\[
d_i \in [0, C_i]
\]
where \( C_i \) is the number of categories for feature \( i \).

\textbf{Continuous and integer features}, which are not natively supported by MPS, are transformed via quantization. Integer and monetary features are scaled (e.g., by 100×) and expressed in a numerical base \( B \) (typically 10). Each digit is then encoded into a separate tensor core. This format increases the number of cores per feature but preserves the structure needed for MPS modeling.

The arrangement of features within the MPS chain significantly affects model performance. Features with high cardinality or strong correlations (e.g., \texttt{native-country}) are placed near the center of the chain to improve expressiveness while minimizing bond dimensions.

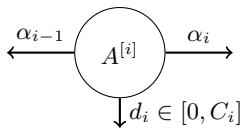
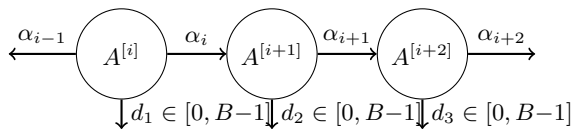
\begin{figure}[ht]
  \centering

  \begin{subfigure}[t]{0.45\textwidth}
    \centering
    \begin{tikzpicture}[node distance=1.2cm, every node/.style={font=\small}]
      \node[draw, circle, minimum size=1.2cm] (feature) at (0,0) {$A^{[i]}$};
      \draw[->, thick] (feature) -- ++(0,-1) node[midway, right] {$d_i \in [0, C_i]$};
      \draw[->, thick] (feature) -- ++(-1.5,0) node[midway, above] {$\alpha_{i-1}$};
      \draw[->, thick] (feature) -- ++(1.5,0) node[midway, above] {$\alpha_i$};
    \end{tikzpicture}
    \caption{Categorical features use a single tensor core with a discrete input index.}
  \end{subfigure}
  \hfill

  \begin{subfigure}[t]{0.5\textwidth}
    \centering
    \begin{tikzpicture}[node distance=1.2cm, every node/.style={font=\small}]
      \node[draw, circle, minimum size=1.2cm] (a1) at (-2,0) {$A^{[i]}$};
      \node[draw, circle, minimum size=1.2cm] (a2) at (0,0) {$A^{[i+1]}$};
      \node[draw, circle, minimum size=1.2cm] (a3) at (2,0) {$A^{[i+2]}$};
      \draw[->, thick] (a1) -- (a2) node[midway, above] {$\alpha_i$};
      \draw[->, thick] (a2) -- (a3) node[midway, above] {$\alpha_{i+1}$};
      \draw[->, thick] (a1) -- ++(0,-1) node[midway, right] {$d_1 \in [0,B{-}1]$};
      \draw[->, thick] (a2) -- ++(0,-1) node[midway, right] {$d_2 \in [0,B{-}1]$};
      \draw[->, thick] (a3) -- ++(0,-1) node[midway, right] {$d_3 \in [0,B{-}1]$};
      \draw[->, thick] (a1) -- ++(-1.5,0) node[midway, above] {$\alpha_{i-1}$};
      \draw[->, thick] (a3) -- ++(1.5,0) node[midway, above] {$\alpha_{i+2}$};
    \end{tikzpicture}
    \caption{Continuous features are digitized in base-\( B \) and encoded over multiple tensor cores.}
  \end{subfigure}

  \caption{Encoding schemes for feeding tabular data into the MPS architecture.}
  \label{fig:feature-encoding}
\end{figure}

\subsection{Gradient Clipping}

Gradient clipping is a critical preparatory step for enforcing differential privacy in stochastic optimization, but it does not ensure privacy on its own. Its purpose is to bound the contribution of individual data points to the gradient computation, thereby defining a global sensitivity bound $\Delta f$. This bounded sensitivity is required before calibrated noise can be added to achieve formal privacy guarantees.

Formally, for each sample-wise gradient $\nabla \ell(x_i)$ computed during training, we apply clipping based on a predefined $\ell_2$-norm threshold $C$:

\begin{equation}
\tilde{\nabla} \ell(x_i) = \nabla \ell(x_i) \cdot \min\left(1, \frac{C}{|\nabla \ell(x_i)|_2}\right),
\end{equation}

\noindent where $\tilde{\nabla} \ell(x_i)$ is the clipped gradient. This operation restricts the maximum influence that any single sample can have on the model update, and is essential for enabling valid noise calibration under the differential privacy accounting framework.

\subsection{Noise Injection}

In differential privacy, noise injection is essential for protecting sensitive data while maintaining analytical utility. We compare two of the most common approaches, Gaussian and Laplacian noise injection, each suited to different data characteristics and privacy requirements.

Gaussian noise adds randomness drawn from a normal distribution to gradients during stochastic gradient descent (SGD), effectively masking individual data points. It is well-suited for continuous data or when feature distributions are approximately Gaussian. However, its uniform dispersion across dimensions can be suboptimal for high-dimensional or sparse datasets.

Laplacian noise, by contrast, is sampled from a distribution with heavier tails, which helps preserve sparsity and structural features in the data. This makes it more effective for datasets with outliers or sparse representations. Laplacian noise is also computationally efficient, particularly in high-dimensional settings, and can offer stronger theoretical privacy guarantees under certain conditions.

In our experiments, we compare Gaussian and Laplacian mechanisms, observing trade-offs between privacy strength and downstream model utility.

We adopt the moment accounting method from Abadi et al.~\cite{Abadi_2016} to track cumulative privacy loss during SGD. This approach rigorously quantifies the privacy-utility trade-off but requires careful calibration of the noise distribution and its interaction with model parameters. Noise is added at each gradient step, with magnitude calibrated based on dataset sampling ratio and privacy parameters according to:

\begin{equation}
\label{eq:MA}
\sigma \geq c \cdot q \cdot \frac{\sqrt{T \log \left( \frac{1}{\delta} \right)}}{\epsilon} .
\end{equation}

\noindent Where:
\begin{itemize}
\item $\sigma$: noise standard deviation.
\item q: sampling ratio (mini-batch size / dataset size).
\item T: number of gradient steps.
\item $\epsilon$ and $\delta$: DP parameters.
\item $\Delta f$: global sensitivity bound.
\end{itemize}

In Laplacian implementation, we applied noise ($\sigma$) for every update using:

\begin{equation}
\label{eq: laplacian}
\sigma = \frac{\Delta f}{\epsilon}
\end{equation}








\subsection{Training Setup and Hyperparameters}

The model is trained via mini-batch stochastic gradient descent (SGD), with hyperparameters designed to minimize privacy cost:

\begin{itemize}
    \item \textbf{Number of batches}: distinct mini-batches per epoch.
    \item \textbf{Descent steps}: SGD updates per epoch.
\end{itemize}

To reduce the total privacy budget \( \epsilon \), we ensure that the number of descent steps is less than or equal to the number of batches, thereby minimizing the number of noisy updates.

\subsection{Evaluation Metrics}

To evaluate model performance, we use both \textbf{fidelity metrics} and \textbf{quality metrics}.

\subsubsection{Fidelity Metrics}

Fidelity scores are computed using the SDMetrics library~\cite{sdmetrics}, which provides standardized evaluation tools for synthetic tabular data. Each metric is evaluated over 10 synthetic samples per configuration. These metrics assess the statistical similarity and distance between synthetic and real data columns:

\textbf{Categorical feature metrics}:
\begin{itemize}
    \item \textbf{Category Coverage} - ensures all categories from the original data appear in the synthetic dataset.
    \item \textbf{Total Variation Distance} - compares marginal distributions.
    \item \textbf{Chi-Square Test} - evaluates statistical significance between observed and expected frequencies.
    \item \textbf{Contingency Similarity} - preserves joint distributions between categorical feature pairs.
\end{itemize}

\textbf{Continuous/Integer feature metrics}:
\begin{itemize}
    \item \textbf{Boundary Adherence} - validates min/max range consistency.
    \item \textbf{Range Coverage} - checks completeness of coverage over the data domain.
    \item \textbf{Kolmogorov-Smirnov (KS) Test} - compares full distributions between synthetic and real values.
\end{itemize}

Each metric is reported as a percentage of similarity (0-100\%).

\subsubsection{Downstream Utility and Quality Metrics}

Beyond statistical similarity, a crucial aspect of synthetic data evaluation is its ability to support real-world machine learning tasks. To assess this, we adopt a \textbf{downstream utility} approach: we train classification models on synthetic data and evaluate their performance on a real, held-out test set. This evaluates whether synthetic samples preserve not only marginal and joint distributions, but also task-relevant signal and feature-target relationships.

We use four widely adopted classifiers from Scikit-learn:
\begin{itemize}
    \item K-Nearest Neighbors (KNN)
    \item Random Forest
    \item Gradient Boosting
    \item Support Vector Machine (SVM)
\end{itemize}

All classifiers are tasked with predicting a binary income label, and their performance is measured using the \textbf{F1-score}:
\begin{equation}
\text{F1} = \frac{2 \cdot \text{TP}}{2 \cdot \text{TP} + \text{FP} + \text{FN}} .
\end{equation}

The real dataset is partitioned into 60\% training, 20\% validation, and 20\% test. For the synthetic data, we apply an 80\%-20\% split for training and validation, while maintaining identical sample sizes to ensure fair model comparison. All models are trained separately on both real and synthetic domains and tested on the same real test data to control for sampling effects.

\paragraph{Interpretation.} High F1 scores for classifiers trained on synthetic data suggest that the generated samples retain sufficient semantic and structural fidelity for predictive modeling. In other words, utility metrics complement fidelity by validating that synthetic data is not only statistically similar, but also functionally informative for machine learning tasks.

\paragraph{Note on KNN.} K-Nearest Neighbors is often used as a proxy for data similarity due to its reliance on sample distances. While in this work it is applied in a supervised classification setting, it implicitly serves as a distance-based validation tool. A fully unsupervised KNN evaluation (e.g., for kNN-based coverage or memorization analysis) is nontrivial due to the mixed presence of categorical and continuous features, which complicate the definition of a unified metric space.

\section{Results}
\label{sec:results}

We evaluate the performance of the MPS-based generative model under three configurations, each reflecting a different stage of privacy mechanism integration:

\begin{enumerate}
    \item \textbf{No Privacy} - MPS is trained without gradient clipping or noise injection. This baseline setup evaluates the model's optimal fidelity and utility without privacy constraints.
    
    \item \textbf{Clipping Only (Sensitivity Control)} - Gradient clipping is applied to bound the sensitivity of individual data contributions. While this prepares the model for differential privacy, it does not in itself ensure any formal privacy guarantees.
    
    \item \textbf{Differential Privacy (DP-MPS)} - Full differential privacy is enforced through the combination of gradient clipping and calibrated noise injection, with results evaluated across multiple $\epsilon$ values.
\end{enumerate}

\subsection{No Privacy - Unperturbed Synthetic Data}
\label{sec:results-no-privacy}

We begin by assessing the performance of the MPS model when no privacy mechanisms are applied. This configuration serves as a reference for upper-bound fidelity and downstream task utility.

\begin{figure}[htbp]
  \centering
  \includegraphics[width=0.48\textwidth]{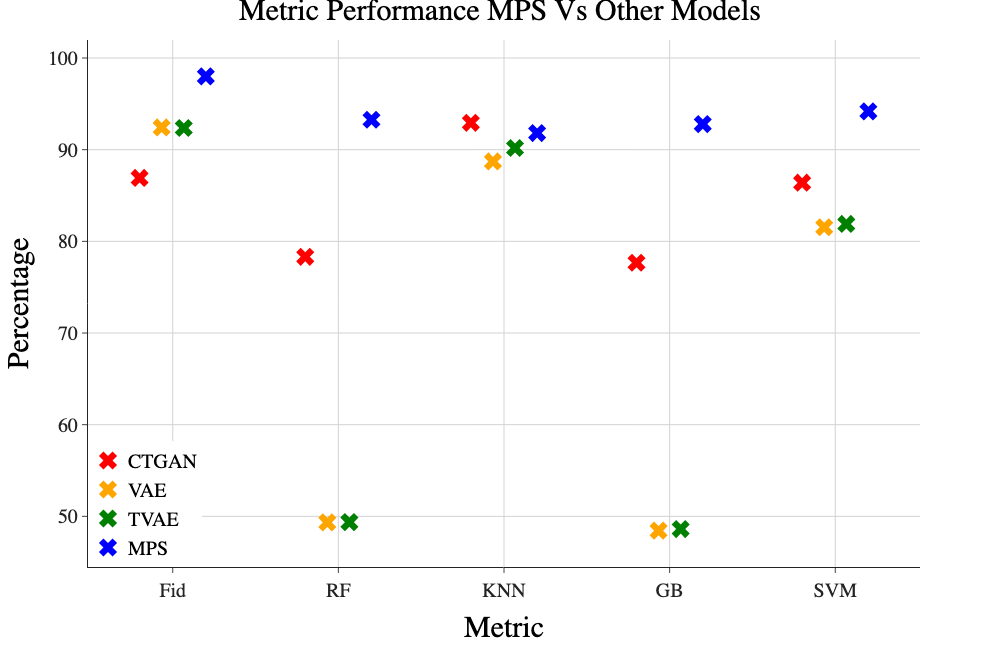}
  \caption{Metric performance for MPS vs other synthetic data models, and the real data.}
  \label{fig:metric_performance_4}
\end{figure}

Figure~\ref{fig:metric_performance_4} shows the overall comparison between MPS and classical baselines (CTGAN, VAE, and TVAE) across fidelity and classification quality. MPS outperforms all baselines, with fidelity scores near 100\% and classifier F1 scores closely matching the real data performance. 

The results indicate that after MPS, CTGAN is the best performing benchmark approach across most quality checks. However, it slightly lags behind in the fidelity test compared to VAE and TVAE. Desipte this, whilst VAE and TVAE exhibit good performance levels when assessed with KNN and SVM, they degrade on the RF and GB metrics. The distinct strengths across different evaluation metrics implies how different structures in the data are preserved across methodologies. In contrast, the MPS synthetic data outperform all benchmark models in all metrics, closely resembling the real data metrics. This demonstrates that the MPS is able to closely preserve all the important relationships within the data that traditional synthetic data algorithms struggle to consistently preserve. 

To provide a comprehensive view of fidelity performance, we break down the results across individual metrics. Table~\ref{tab:fidelity_stats} presents the mean and standard deviation for each fidelity metric, computed over ten independent runs. All metrics are intrinsically defined within the range \([0, 1]\), where values closer to 1 indicate higher similarity between synthetic and real data distributions. 


\begin{table}[htbp]
\centering
\begin{tabular}{|l|c|c|}
\hline
\textbf{Fidelity Metric} & \textbf{Mean} & \textbf{Standard Deviation} \\
\hline
Category Coverage & 0.9979 & 0.0011 \\
Total Variation & 0.9966 & 0.0004 \\
Chi-Square & 0.9993 & 0.0003 \\
Contingency Similarity & 0.8585 & 0.0007 \\
Boundary Adherence & 0.9992 & 0.0001 \\
Range Coverage & 0.9889 & 0.0123 \\
Kolmogorov-Smirnov & 0.9969 & 0.0004 \\
\hline
\end{tabular}
\caption{Disaggregated fidelity metrics for MPS across 10 random seeds. Results show high stability and accuracy.}
\label{tab:fidelity_stats}
\end{table}

\begin{figure}[htbp]
  \centering
  \includegraphics[width=0.48\textwidth]{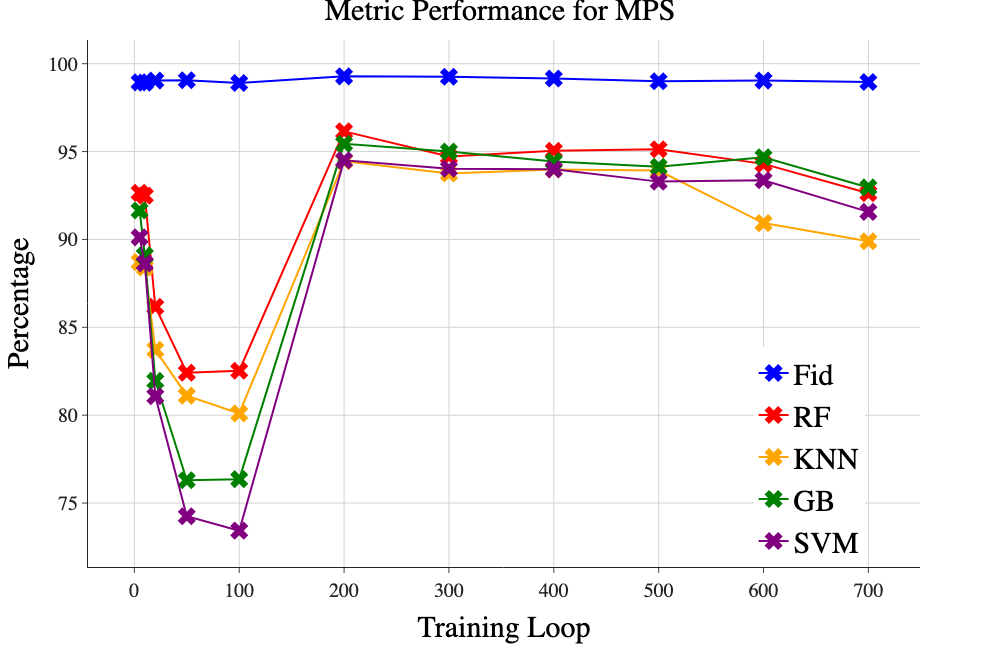}
  \caption{Metric performance over training loops in MPS. Fidelity stabilizes early, while utility improves up to 200 loops.}
  \label{fig:metric_performance_3}
\end{figure}

The evolution of performance across training loops is shown in Figure~\ref{fig:metric_performance_3}. Fidelity stabilizes after relatively few iterations, while downstream utility (F1 score) improves with training, peaking around 200 loops. This suggests that MPS captures basic statistical structure early, and later iterations help refine more complex dependencies.

These results highlight the ability of the MPS model to generate synthetic data with high fidelity and strong utility performance in the absence of privacy constraints. The model consistently captures key statistical properties across different trials and training loops, making it a strong candidate for privacy-aware extensions evaluated in the following sections.

\subsection{Gradient Clipping - Stability Without Noise}
\label{sec:clipped-mps}

Before introducing differential privacy guarantees, we evaluate the effect of gradient clipping alone. Although clipping does not ensure privacy by itself, it is a prerequisite for differential privacy, as it bounds the sensitivity of gradients before noise injection.

\begin{figure}[htbp]
  \centering
  \includegraphics[width=0.48\textwidth]{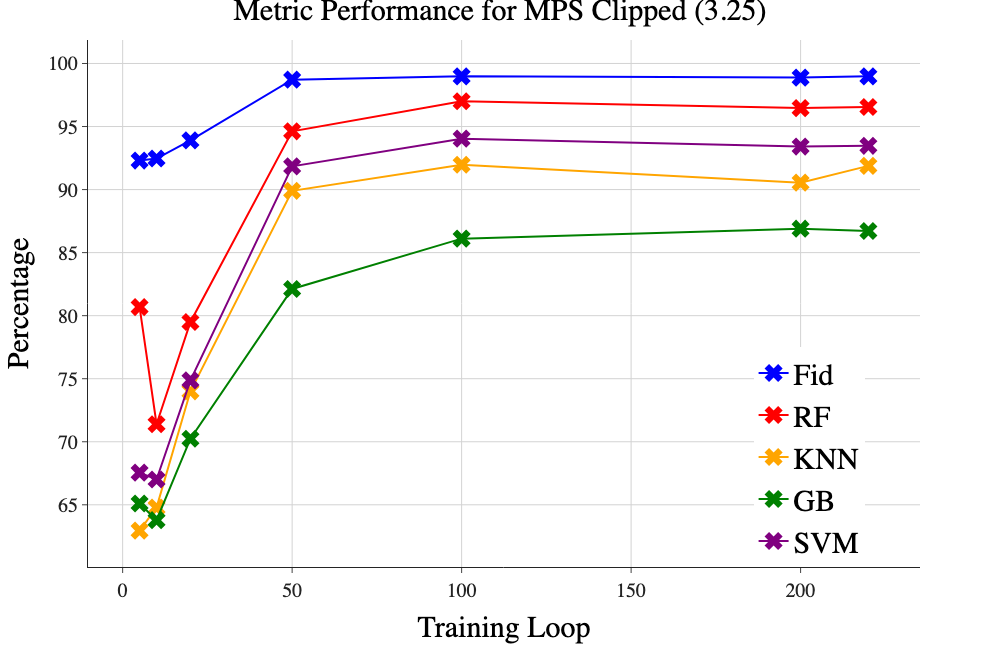}
  \caption{Metric performance over training loops for Clipped MPS. Convergence is reached with fewer steps.}
  \label{fig:clip_metric_performance_6}
\end{figure}

This intermediate configuration, referred to as \emph{Clipped MPS}, enables the assessment of training stability and performance in isolation from noise. Clipping prevents extreme gradients from dominating updates, and empirically helps mitigate training instabilities such as gradient explosion or vanishing gradients. 

Figure~\ref{fig:clip_metric_performance_6} shows the progression of model performance over training loops using clipping. Compared to the unclipped model, stabilization is observed at 100 loops compared to 200 for the unclipped model. This behavior suggests improved convergence dynamics. Additional convergence plots comparing clipped models at 100 and 200 loops are provided in Appendix~\ref{app:clipping-analysis}, see Figure~\ref{fig:appendix-clipping-plot} for details.

\begin{figure}[htbp]
  \centering
  \includegraphics[width=0.48\textwidth]{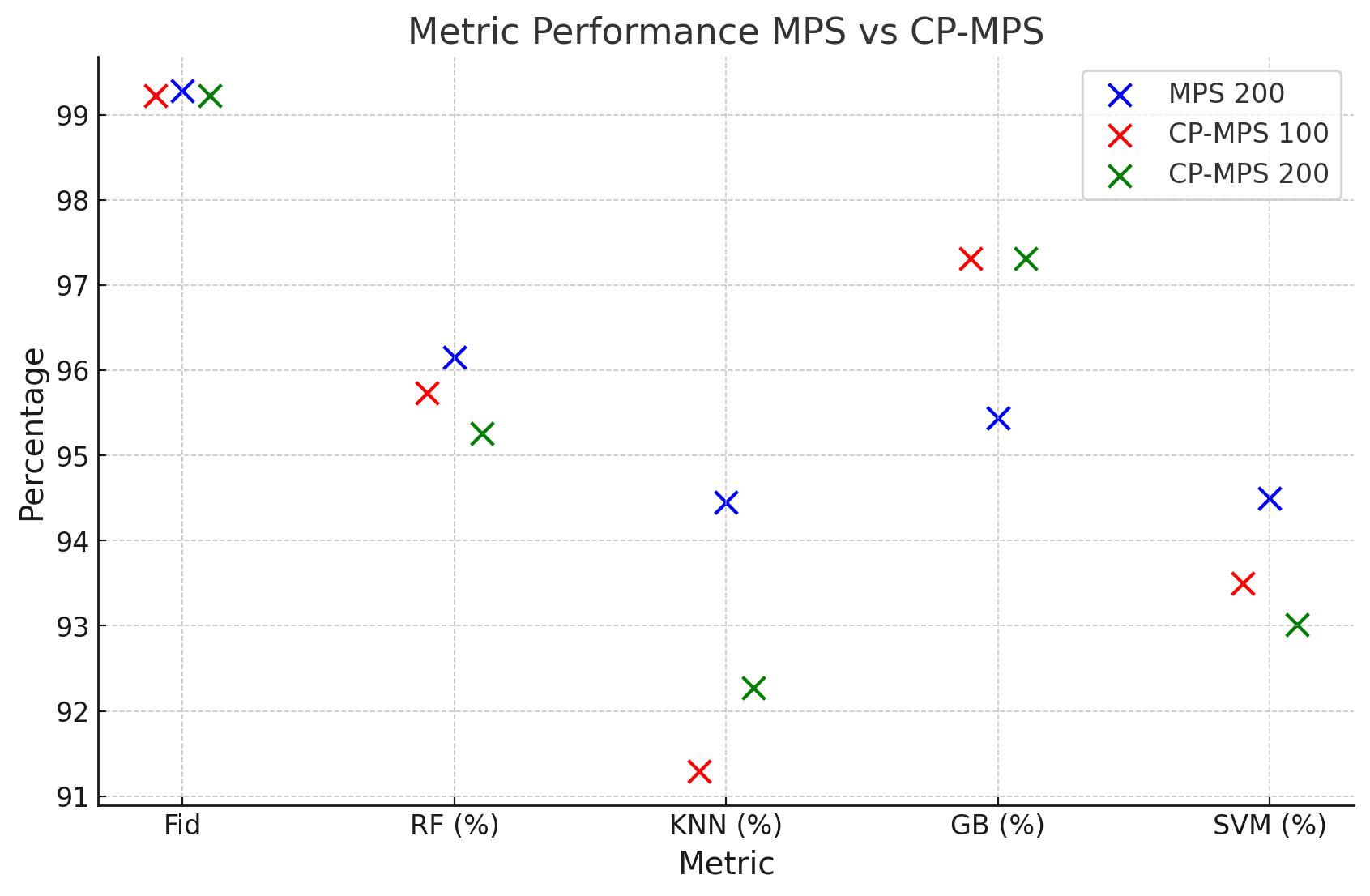}
  \caption{Comparison between Clipped MPS (100 loops) and baseline MPS (200 loops).}
  \label{fig:metric_performance_6.1}
\end{figure}

To compare Clipped MPS against the baseline (unclipped) MPS, we evaluate both at their respective best-performing training loops (Clipped at 100, unclipped MPS at 200). Figure~\ref{fig:metric_performance_6.1} presents this comparison across fidelity and downstream classification tasks.

For all metrics the Clipped MPS model shows high performance with metric scores in the 90\% range for all tasks. The Clipped MPS model achieves comparable fidelity and Random Forest performance, and shows improved results in Gradient Boosting. However, it underperforms in KNN and SVM classifiers. Graident is effective for limiting the individual contributions of outliers which could be contributing to the under-performance for KNNs; this is an indicator of some basic privacy injection showing that distances in the data have been distorted. These differences highlight the varying sensitivity of classification algorithms to the learned data distribution. 
The next step integrates calibrated noise injection into the clipped model to achieve formal privacy guarantees.

\subsection{Noise Injection Differential Privacy - Privacy Guarantees} 
\label{sec:DP-results}

To implement full differential privacy (DP) guarantees in the MPS generative framework, we apply noise injection mechanisms during training. Both Gaussian and Laplacian noise distributions are tested, with calibration based on the moment accounting formulation discussed in Section~\ref{sec:experimental_setup}. All experiments use 100 training loops.

\begin{figure}[htbp]
  \centering
  \includegraphics[width=0.48\textwidth]{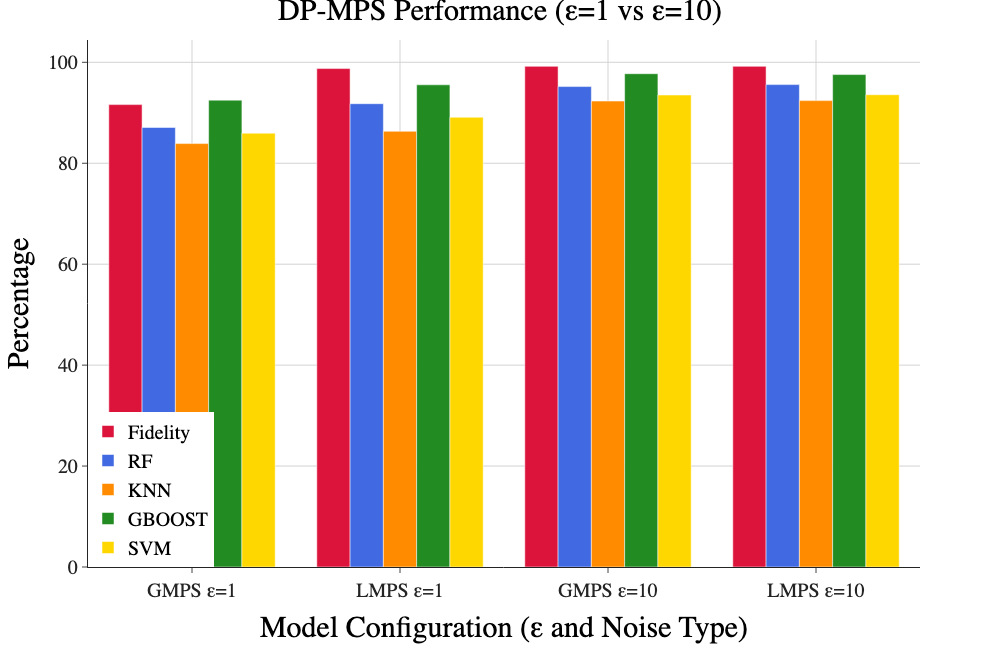}
  \caption{Metric performance for DP-MPS with Gaussian Vs Laplacian noise injection across multiple $\epsilon$ values.}
  \label{fig:dp_metric_performance_6}
\end{figure}

Figure~\ref{fig:dp_metric_performance_6} reports the performance of the DP-MPS model trained with Gaussian and Laplacian noise injection (GMPS and LMPS respectively), for different levels of the privacy parameter $\epsilon$. This method omits gradient step accounting and instead calibrates noise directly per update using global sensitivity bounds. As expected, higher $\epsilon$ (i.e., lower noise) yields improved performance across all metrics, we also see that due to the presence of outliers in the data the Laplacian method performs better.


\begin{figure}[htbp]
  \centering
  \includegraphics[width=0.48\textwidth]{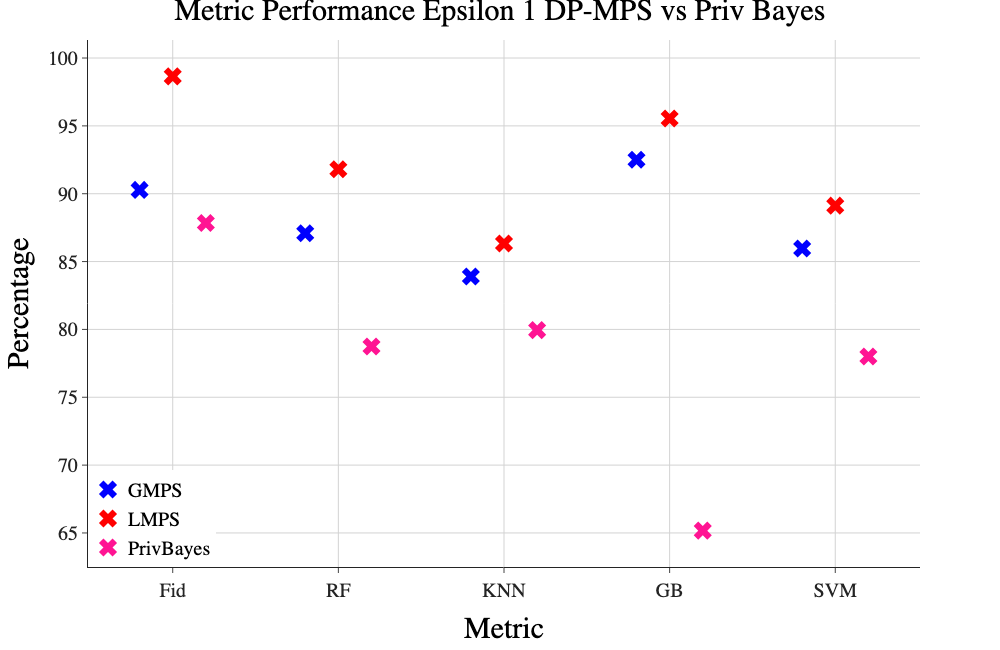}
  \caption{Performance comparison at $\epsilon = 1$ between DP-MPS and PrivBayes.}
  \label{fig:metric_performance_11}
\end{figure}

\begin{figure}[htbp]
  \centering
  \includegraphics[width=0.48\textwidth]{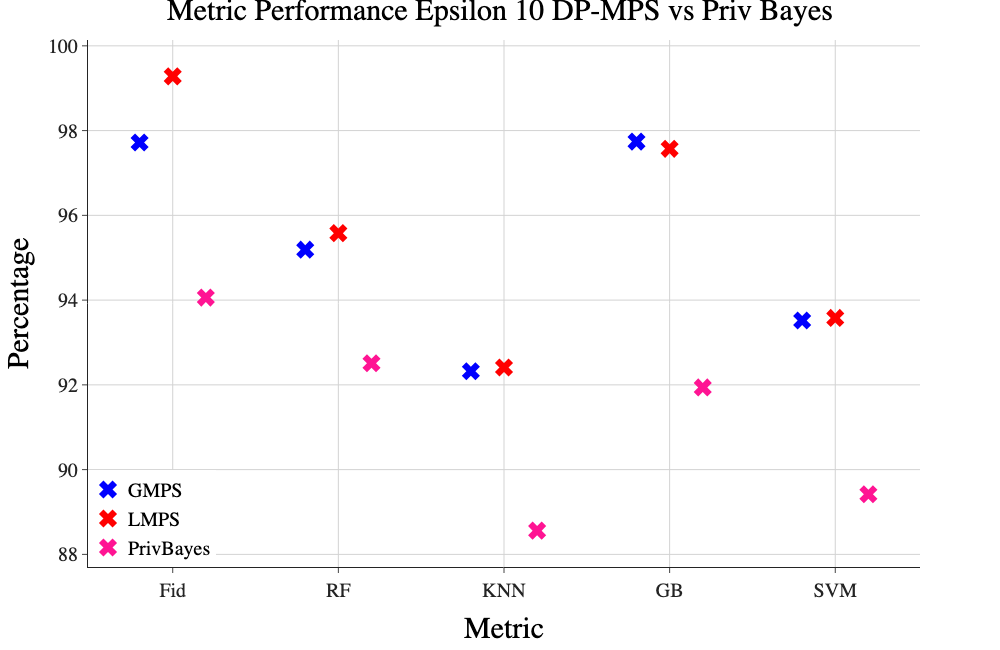}
  \caption{Performance comparison at $\epsilon = 10$ between DP-MPS and PrivBayes.}
  \label{fig:metric_performance_8}
\end{figure}

To assess the comparative behavior of DP-MPS against a classical DP baseline, we include performance plots for $\epsilon = 1$ and $\epsilon = 10$ contrasting DP-MPS and PrivBayes in Figures \ref{fig:metric_performance_11} and \ref{fig:metric_performance_8} respectively. These configurations illustrate the effect of strict and relaxed privacy constraints. We see that for the same $\epsilon$ privacy budget the DP-MPS synethetic data provides significant improvements over all the assessed metrics. In particular, on downstream tasks we see that DP-MPS provides superior synthetic data. These results demonstrate that DP-MPS can provide a valuable tool to improve the analysis of sensitive data. 

\section{Conclusion and outlook}
\label{sec:conclusion}
In this work we demonstrate the significant potential of Matrix Product States (MPS) in the field of synthetic data generation. By employing MPS, we have demonstrated its superior performance in generating high-fidelity synthetic data compared to established models like CTGAN, VAE, and PrivBayes for data privacy. 

Our main contribution is the integration of differential privacy into the MPS model. This addition not only enhances privacy protections but also ensures that the generated synthetic data remains practical and reliable for diverse applications. Our experimental results show that MPS outperforms CTGAN and PrivBayes, especially in scenarios that demand stringent privacy preservation, achieving an overall utility improvement of more than 10{\%} at standard privacy levels and over 8{\%} at the highest privacy settings.

The ability of MPS to generate high-quality, privacy-preserving synthetic data paves the way for more secure data sharing and analysis, addressing the critical challenges of data scarcity and privacy constraints. This can be impactful in fields such as medical machine learning learning which rely on high-quality private data to protect patient confidentiality. 

In conclusion, this work demonstrates that quantum-inspired machine learning using MPS offers a compelling alternative for synthetic data generation. MPS has shown significant improvements in data fidelity, privacy preservation, and utility metrics, suggesting its potential as a powerful tool in this field.

\section{Acknowledgments}
\label{sec:acknowledgments}
\vspace{-5pt}
This work is a collaboration between Bundesdruckerei Group and Multiverse Computing. We would like to acknowledge and thank the technical teams and field engineers both at Bundesdruckerei Group and Multiverse Computing for extremely helpful discussions.

\bibliographystyle{apsrev4-1} 
\bibliography{QCAPaper} 
\appendix
\section{Additional Results for Clipping Experiments}
\label{app:clipping-analysis}

\begin{figure}[htbp]
  \centering
  \includegraphics[width=0.48\textwidth]{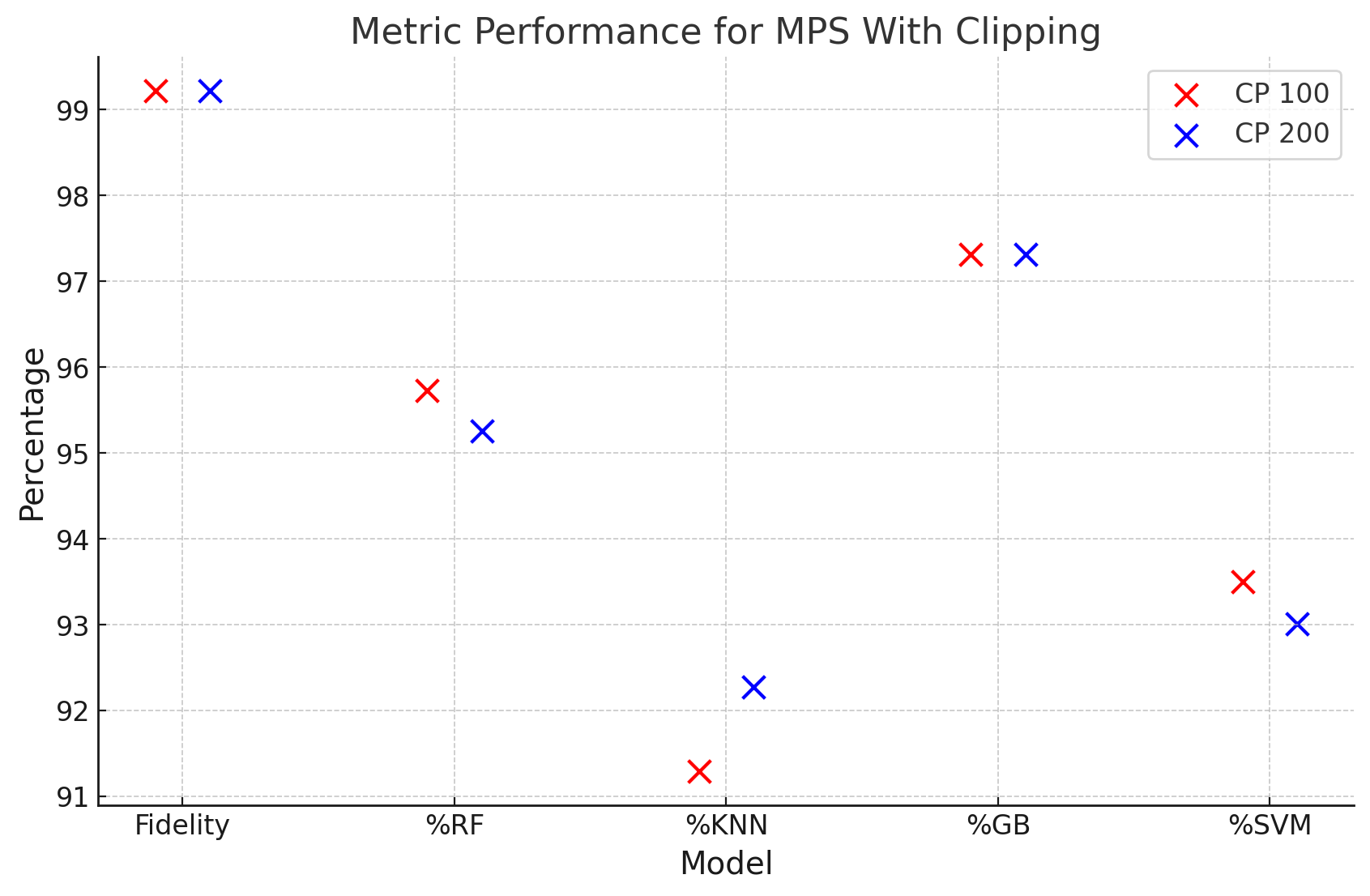}
  \caption{Comparison of Clipped MPS performance at 100 and 200 training loops.}
  \label{fig:appendix-clipping-plot}
\end{figure}

\subsection{PrivBayes Performance at Different Privacy Levels}
\label{sec:appendix_privbayes}

\begin{figure}[htbp]
  \centering
  \includegraphics[width=0.48\textwidth]{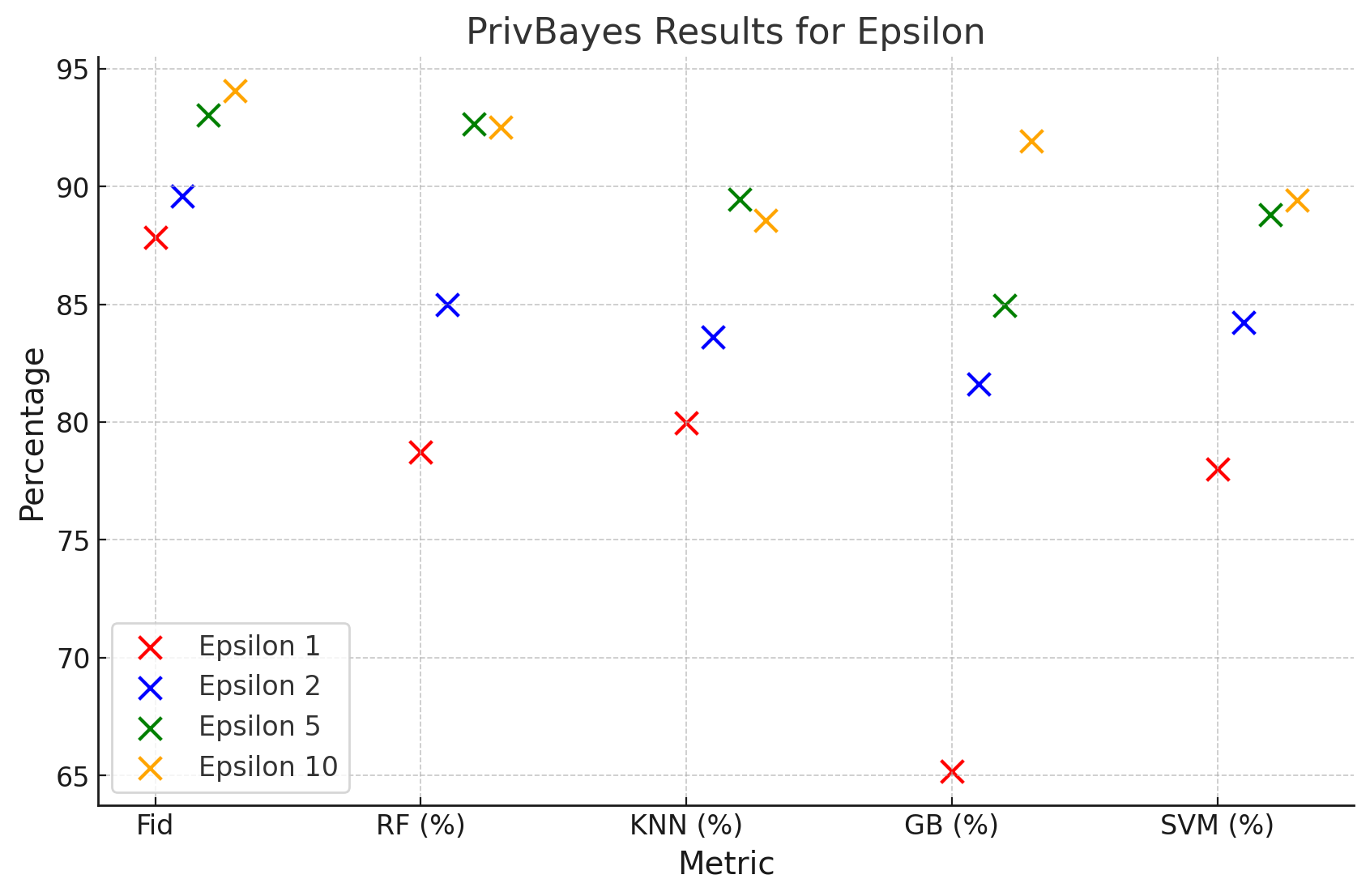}
  \caption{Metric performance for PrivBayes at varying privacy levels $\epsilon \in \{1, 2, 5, 10\}$.}
  \label{fig:metric_performance_5.5}
\end{figure}

To provide a baseline comparison, Figure~\ref{fig:metric_performance_5.5} reports the performance of the PrivBayes model under four different privacy budgets ($\epsilon \in \{1, 2, 5, 10\}$). As expected, higher values of $\epsilon$ correspond to improved fidelity and downstream classification performance, due to reduced noise injection. This figure serves as a reference point for interpreting the results of the DP-MPS model introduced in this work.

\subsection{Additional Comparison: DP-MPS vs. PrivBayes at Intermediate $\epsilon$ Values}
\label{sec:appendix_comparison_eps5_2}

To supplement the main analysis focused on $\epsilon = 1$ and $\epsilon = 10$, we include additional comparison plots for intermediate values $\epsilon = 2$ and $\epsilon = 5$. These figures illustrate how both fidelity and downstream classification quality improve for the DP-MPS model relative to PrivBayes as the privacy budget increases. The patterns follow expected trends, with DP-MPS consistently outperforming PrivBayes across all classifiers and metrics.

\begin{figure}[htbp]
  \centering
  \includegraphics[width=0.48\textwidth]{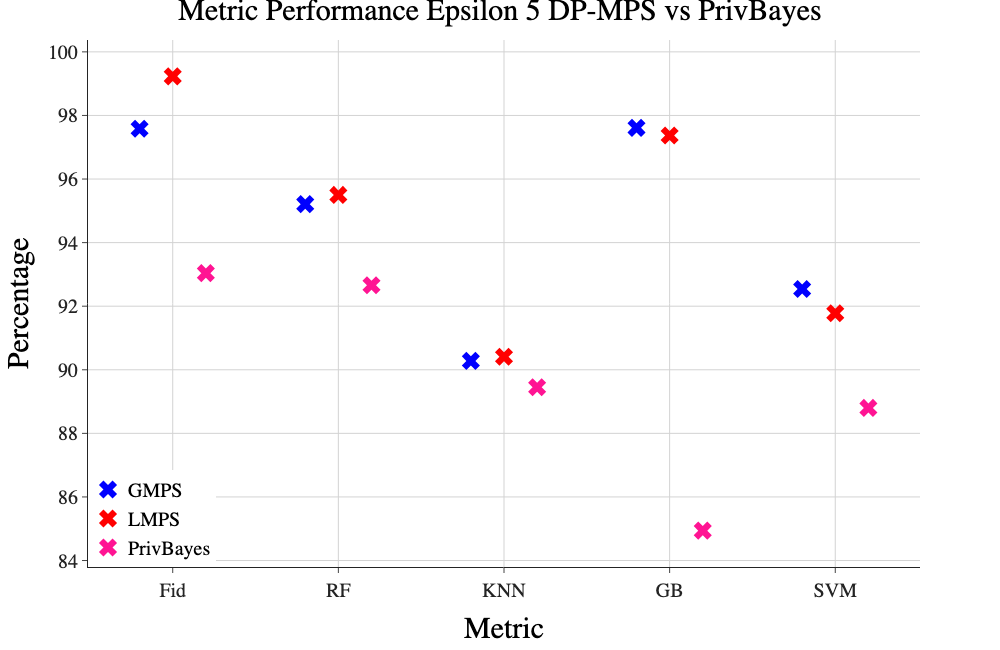}
  \caption{Metric performance for DP-MPS vs PrivBayes at $\epsilon = 5$.}
  \label{fig:metric_performance_9}
\end{figure}

\begin{figure}[htbp]
  \centering
  \includegraphics[width=0.48\textwidth]{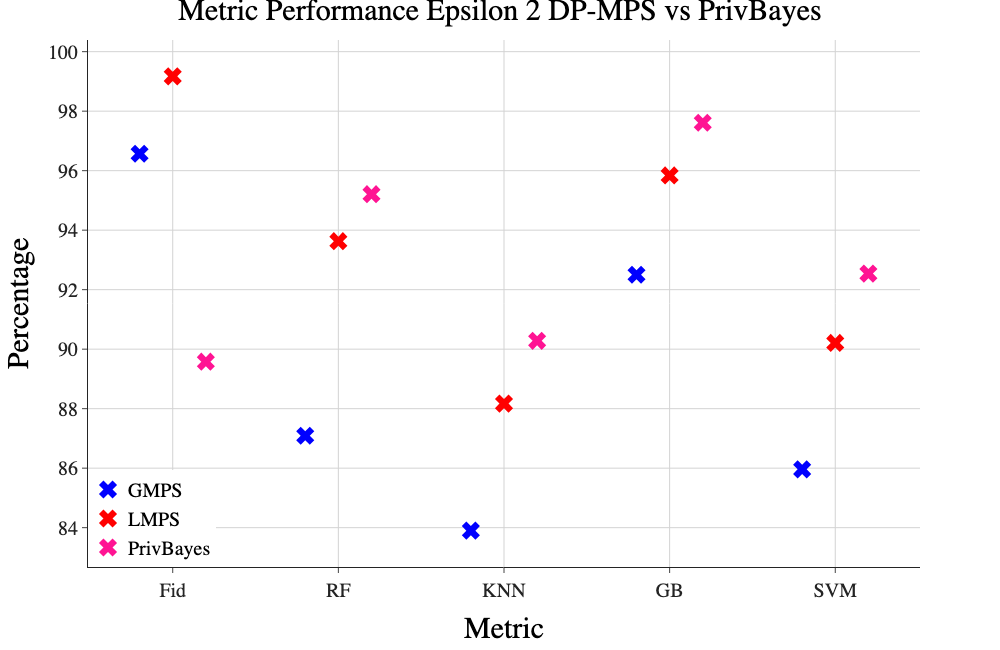}
  \caption{Metric performance for DP-MPS vs PrivBayes at $\epsilon = 2$.}
  \label{fig:metric_performance_10}
\end{figure}

\begin{figure}[htbp]
  \centering
  \includegraphics[width=0.48\textwidth]{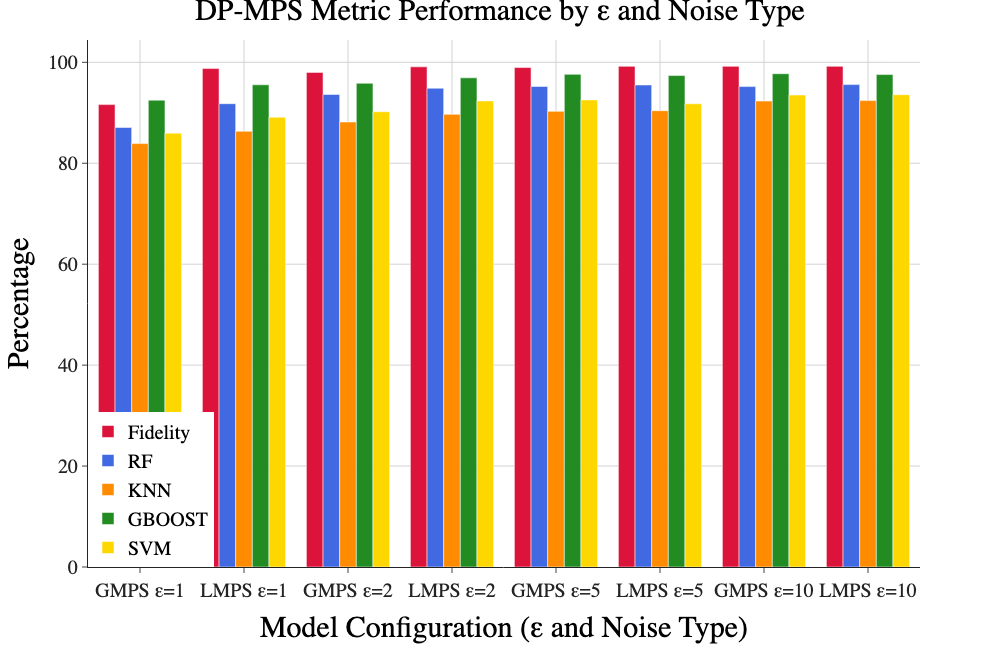}
  \caption{Metric performance for DP-MPS with Gaussian Vs Laplacian noise injection across multiple $\epsilon$ values.}
  \label{fig:metric_performance_6}
\end{figure}

\end{document}